\documentclass[10pt]{article} 
\usepackage[preprint]{tmlr}

\usepackage{amsmath,amsfonts,bm}

\def\eqref#1{equation~\ref{#1}}

\def\1{\bm{1}}

\DeclareMathAlphabet{\mathsfit}{\encodingdefault}{\sfdefault}{m}{sl}
\SetMathAlphabet{\mathsfit}{bold}{\encodingdefault}{\sfdefault}{bx}{n}

\usepackage{hyperref}
\usepackage{url}
\usepackage{float}
\usepackage{graphicx}

\title{On Cross-Validation for Hyperparameter Optimization of Deep
  Learning Image Classifiers}

\author{\name Ljubomir Buturovi\'c \email ljubomir@gmail.com \\
      \addr East Palo Alto, California}

\def\month{MM}  
\def\year{YYYY} 
\def\openreview{\url{https://openreview.net/forum?id=XXXX}} 

\begin{document}

\maketitle

\begin{abstract}

Hyperparameter optimization (HPO) can materially affect the
performance of deep learning (DL) image classifiers, but there is
little empirical guidance on how to derive the validation signal that
drives it, especially for the small sample sizes common in fields such
as medical imaging. We compared three HPO protocols in terms of {\em
  absolute performance-estimation error} (AEE; the absolute difference
between the winning configuration's validation AUROC and its test
AUROC): fixed holdout ({\em {\bf F}}), reshuffled holdout ({\em {\bf
    R}}), and 5-fold cross-validation ({\em {\bf C}}). The search
space, sampler, training procedure, architecture, and test set were
held identical across protocols. We evaluated the protocols on three
public datasets spanning two regimes: binary medical imaging (RSNA
pneumonia radiographs and binarized HAM10000 skin lesions) and
200-class natural imaging (Tiny ImageNet), across a range of
development set sizes $n$ and two backbones (ResNet-18 on all
datasets, Vision Transformer (ViT-S/16) on RSNA). On the medical
datasets, every point estimate favored cross-validation over both
holdout protocols, with reductions in AEE largest at small sample
sizes and diminishing as $n$ increased. This pattern remained robust
under conservative family-wise adjustment.  On Tiny ImageNet, AEE was
negligible under all three protocols. Test AUROC was generally similar
among protocols. Fixed holdout had lower mean AEE than reshuffled
holdout in 11 of 12 medical conditions, although this secondary
finding was less uniformly supported. For small-sample medical image
classification, we recommend cross-validation-based HPO when
computational resources permit because it trades additional
computation for a more reliable development-time estimate of
subsequent test performance.

\end{abstract}
  
\section{Introduction}

Hyperparameter optimization (HPO) can significantly improve the
performance of predictive models in machine learning (ML)
\citep{probst2019tunability, bischl2023hpo, feurer2019hpo} and deep
learning (DL) \citep{schmidt2021crowdedvalley}. The HPO methods used
in ML and DL differ.  In classical, non-DL ML, HPO commonly evaluates
quality of a hyperparameter configuration (HC) by using
cross-validation (CV) estimates of the relevant predictive performance
statistic, such as AUROC~\citep{krstajic2014cross}. The HC space is
searched by specialized algorithms such as random
search~\citep{bergstra2012random} or Hyperband~\citep{hyperband}. In
DL, when HPO is used, it typically evaluates a classifier trained
using an HC by applying it to a fixed validation set
\citep{bischl2023hpo} and recording the performance statistics. In
both approaches, the final model utilizes the HC that yielded the best
performance during optimization.

CV is generally considered computationally prohibitive and unnecessary
for large image datasets. However, for smaller datasets, which are
common in important applications such as classification of medical
images, CV may be feasible, yet there is little published guidance on
its merits for HPO.  In particular, to our knowledge, there are no
empirical studies systematically comparing CV and fixed validation set
approaches for DL image classification HPO.

In this paper, we aim to fill the gap by comparing the performance of
DL image classifiers tuned using held-out validation and CV
approaches. The goal is to help practitioners and researchers make an
informed choice regarding balancing the tuning bias and computational
cost of the HPO methods.

\section{Methods}
\label{methods}

\subsection{Overview}

The primary goal was comparison of leading hyperparameter optimization
(HPO) protocols for tuning of deep image classifiers: a fixed holdout
({\em {\bf F}}) and 5-fold cross-validation ({\em {\bf C}}). We also
included a recent proposal of reshuffled holdout ({\em {\bf R}})
\citep{Nagler2024}. Thus, we compared the three methods. To isolate
the effect of the evaluation method, the search space, the HPO sampler
(Optuna's Tree-structured Parzen Estimator (TPE) sampler; no
early-stopping scheduler was was used), training procedure,
architecture, and test data were held identical across all
approaches. The three protocols differ in how the validation signal
driving the HPO is derived:

\begin{itemize}

\item {\bf {\em F}}: A single train/validation split was drawn once
  and reused to score every HPO trial. 

\item {\bf {\em R}}: A fresh random train/validation split was drawn
  for each trial, following Nagler et al. (2024).

\item {\bf {\em C}}: Each trial was scored by the mean validation
  metric over a fixed set of 5 folds.

\end{itemize}

The combination of training and validation data together was referred
to as the {\em training pool}.

Under all three protocols, the winning configuration was retrained
once on the full training pool and evaluated on a held-out test
set. Note that the goal was not achieving SOTA performance on the test
data, but rigorously comparing the HPO methods.

We intentionally held the number of HPO trials, rather than total
compute, constant. Five-fold CV therefore used approximately five
times as many model fits per trial as either holdout protocol. This
additional computation is an inherent feature of {\em {\bf C}} and is
the mechanism by which it can reduce variance in the validation
signal. Our comparison evaluates whether this compute-for-reliability
trade is beneficial; it is not a comparison under equal computational
budgets.

\subsection{Datasets}

We chose three image classification datasets based on the following
criteria:

\begin{itemize}

\item publicly available image datasets
  
\item prioritize medical datasets, due to their importance and the
  frequent occurrence of small sample sizes

\item include one non-medical dataset, to generalize conclusions
  outside of the medical domain

\item binary and multi-class problems

\item non-saturated performance using off-the-shelf, non-customized
  CNN architectures (i.e., the accuracy sufficiently far from 100\% to
  enable reliable ranking of the methods)

\item sufficient number of images available to accurately estimate
  performance of the final tuned model using independent test set data
  
\end{itemize}

The search yielded three datasets (named {\em RSNA}, {\em BHAM} and
{\em TIN}) spanning two regimes: binary medical imaging and many-class
natural imaging. RSNA is the RSNA Pneumonia Detection Challenge
(\url{https://kaggle.com/competitions/rsna-pneumonia-detection-challenge};
binary chest radiograph classification, pneumonia versus not;
pneumonia prevalence: ~23.0\%). BHAM is HAM10000
(\citep{tschandl2018ham10000}; Harvard Dataverse,
\url{doi:10.7910/DVN/DBW86T}) binarized into malignant or premalignant
({akiec, bcc, mel}) versus benign ({bkl, df, nv, vasc}) skin lesions
(positive-class prevalence: 19.5\%). Train/test splits are
lesion-disjoint, produced by stratified group k-fold on lesion\_id so
that no lesion's images appear in both partitions. TIN is Tiny
ImageNet (\url{http://cs231n.stanford.edu/tiny-imagenet-200.zip};
200-class natural image classification).

For each dataset, we downloaded raw images from the Internet and
partitioned into a large held-out test set and a smaller training
pool, consistent with the goals of this research.  Within each
subsample the development data was used either as an 80/20
train/validation holdout (methods {\bf {\em F}} and {\bf {\em R}}) or
in its entirety under 5-fold cross-validation (method {\bf {\em C}}),
while the test set was common to all methods. RSNA DICOM images were
stratified into patient-disjoint train/test split over the unique
patientIds (test\_size=0.8), deliberately reserving the larger 80\%
(21,348 images) as the test set and leaving the remaining 5,336 as the
training pool. We drew five independent class-stratified subsamples of
size $n \in \{100, 300, 1000, 3000\}$ from that pool.  For BHAM, we
constructed a binary dataset from the original HAM10000 images and
metadata and partitioned it into development and test sets using
stratified group splitting on \texttt{lesion\_id}, so that no lesion
contributed images to both partitions. The resulting test set
contained 2,003 images from 1,494 lesions, and the remaining 8,012
images from 5,976 lesions formed the development pool. We drew five
stratified subsamples of $n \in \{100,300,1000,3000\}$ from that pool.
TIN used its canonical ~100K-image training pool and repurposed the
official validation set as the test set. We drew five stratified
200-way subsamples of $n \in \{2000, 5000, 10000\}$ from the training
pool (the smaller sample sizes were not feasible for TIN due to 200
classes).  In every case the subsampling was stratified by class with
a fixed base seed (offset per subsample), so the development and test
partitions were disjoint and reproducible, and the test set was never
used for training or to guide tuning.

The datasets sizes are summarized in Table~\ref{tab:datasets}.

\begin{table}[t]
  \caption{Datasets used in the analyses: subsample sizes $n$ and the size of
    the shared held-out test set.}
  \label{tab:datasets}
  \begin{center}
    \begin{tabular}{cccc}
      \multicolumn{1}{c}{\bf Dataset}  & {\bf Classes} & \multicolumn{1}{c}{\bf $n$ values} & \multicolumn{1}{c}{\bf Test images}
      \\ \hline \\
      RSNA            & 2   & 100 / 300 / 1000 / 3000 & 21{,}348 \\
      BHAM            & 2   & 100 / 300 / 1000 / 3000 & 2{,}003  \\
      Tiny ImageNet   & 200 & 2000 / 5000 / 10000     & 10{,}000 \\
    \end{tabular}
  \end{center}
\end{table}

\subsection{Image pre-processing}

To preserve strict parity across our benchmarking environments, all
raw images were processed into a standardized format before model
exposure. The medical imaging sets required deterministic mapping from
native source format to standard image arrays.

For RSNA, we extracted the raw pixel arrays from the source DICOM
files using pydicom. Images recorded in MONOCHROME1 polarity were
inverted to the MONOCHROME2 convention, so that higher pixel values
correspond to higher density. Each image was then scaled to [0,1] by
per-image min-max normalization, mapped to 8-bit [0,255], and resized
to 224 $\times$ 224 using bilinear interpolation. We applied no
dataset-level intensity normalization, so that no cross-image
statistics entered the stored images. The resulting single-channel
images were replicated across three channels ($R=G=B$) at training
time for compatibility with the pretrained backbone. BHAM and TIN
required no such conversion and were used as distributed.

To ensure a controlled benchmarking environment and prevent data
preprocessing from serving as a confounding variable, we adopted the
standard ImageNet normalization pipeline ($\mu = [0.485, 0.456,
  0.406]$, $\sigma = [0.229, 0.224, 0.225]$) across all three
evaluation datasets.

\subsection{Architecture and training}
\label{Architecture}

The primary backbone was an ImageNet-pretrained ResNet-18, used for
all three datasets, with batch-normalization parameters left
unfrozen. To assess robustness to architecture, we additionally ran
the full {\em {\bf F}}/{\em {\bf R}}/{\em {\bf C}} comparison on RSNA
with an ImageNet-pretrained Vision Transformer (ViT-S/16). The final
classification layer was replaced to match the number of classes and
was the only randomly initialized component. Grayscale inputs were
converted to three channels by replication, and all inputs were
normalized with the standard ImageNet per-channel statistics; this was
the only normalization applied at training time.

The hyperparameter space comprised learning rate, weight decay, label
smoothing, dropout rate, RandAugment magnitude and number of
operations, mixup alpha, cutmix alpha and optimizer (AdamW or SGD);
batch size was fixed to 32. The number of epochs was a tunable
hyperparameter between 10 and 50. Configurations were proposed by the
TPE over 50 trials for medical datasets and over 30 trials for TIN,
for practicality reasons. Augmentation was sampled fresh at every
epoch in all protocols.

\subsection{Validation protocols}

The three protocols were implemented identically except for the
validation signal used to rank HPO trials. {\bf {\em F}} drew a single
80/20 train/validation split per subsample and scored every trial on
that fixed split. {\em {\bf R}} drew a fresh random 80/20 split for
each trial. {\bf {\em C}} partitioned the subsample into 5
class-stratified folds; each trial was scored by the mean validation
metric across folds. For BHAM, all train/validation splits and
cross-validation folds were additionally grouped by
\texttt{lesion\_id}, so that images from the same lesion never
occurred on opposite sides of an internal HPO split. At 10 examples
per class (TIN, $n$ = 2000), 5-fold cross-validation places exactly 2
validation examples per class per fold and trains each fold on the
remaining 8 per class, so the per-fit training budget matches the
80/20 holdout of {\bf {\em F}} and {\bf {\em R}}. After HPO, the
winning configuration under each protocol was retrained once on the
full $n$-image subsample and evaluated on the held-out test set.

\subsection{Factorial design and metrics}

For each dataset, sample size, and architecture, we ran a $5 \times S$
factorial of 5 training subsamples crossed with $S$ random seeds. $S$
was 5 for medical datasets and 3 for TIN, to keep computations
manageable. The number of cells per protocol was therefore 25 for
medical datasets and 15 for TIN. The seed controlled the
classification-head initialization, data ordering, augmentation
sampling, dropout, validation-split derivation, and the HPO
sampler. Within each (subsample, seed) cell the three protocols shared
the same seed, so that they differed only in validation-split
derivation; this pairing supports direct paired comparison of the
protocols. The seed was reused across subsamples, yielding a crossed
rather than nested design.

TIN experiments evaluate the HPO protocols under standard
ImageNet-pretrained transfer learning, not from-scratch learning on a
dataset independent of ImageNet pretraining.

The HPO objective metric was AUROC. For the 200-class TIN we used
one-vs-rest AUROC~\citep{fawcett2006introduction}, macro-averaged
(unweighted mean over classes). We chose AUROC because it is a
well-established scalar metric~\citep{van2025evaluation,
  buturovicML4H2025} that does not require choosing a decision
threshold. For each HPO run we also recorded the winning
configuration's validation score $\mathrm{AUROC}_v$ and defined the
absolute performance-estimation error as
$\mathrm{AEE}=|\mathrm{AUROC}_v-\mathrm{AUROC}_t|$, where
$\mathrm{AUROC}_t$ is the test AUROC of the final model retrained on
the full training pool using the winning hyperparameter configuration.
AEE therefore measures the absolute discrepancy between the
development-time performance estimate and subsequent test
performance.

For each protocol comparison ({\em {\bf F}} versus {\em {\bf R}}, and
{\em {\bf C}} versus {\em {\bf F}} and {\em {\bf R}}), we paired runs
within (subsample, seed) cells and calculated the paired difference in
test AUROC and AEE. Because subsamples and random seeds were crossed
rather than nested, we fitted a crossed random-effects model to these
paired differences, with random intercepts for subsample and
seed~\citep{baayen2008mixed}. The model intercept represented the mean
protocol difference, and 95\% confidence intervals were obtained by
parametric bootstrap with 10,000
replicates~\citep{davison1997bootstrap}. As sensitivity analyses, we
additionally used a two-way cluster bootstrap that independently
resampled subsamples and seeds~\citep{owen2007pigeonhole}, and an
analysis in which differences were first averaged across seeds within
each subsample and a Student-$t$ interval was calculated across the
five subsample means.

The 95\% confidence intervals reported for individual contrasts are
pointwise and are not adjusted for multiplicity. The 24
cross-validation-versus-holdout AEE comparisons across the medical
datasets constituted the primary inferential family. As a multiplicity
sensitivity analysis, we constructed Bonferroni-adjusted simultaneous
confidence intervals controlling the family-wise error rate at 0.05
across these 24 comparisons. The simultaneous intervals were estimated
using at least 100,000 parametric-bootstrap replicates per contrast.

For TIN, we additionally report test top-1 accuracy as a secondary
descriptive metric because it is widely used for this dataset, and it
shows the problem is not saturated. Models and hyperparameters
remained selected using macro-averaged one-vs-rest AUROC; top-1
accuracy played no role in HPO.

\subsection{Computational considerations}

Five-fold cross-validation necessarily requires greater computation
than either holdout protocol. Each HPO trial under {\em {\bf F}} or
{\em {\bf R}} required one model fit, whereas each trial under {\em
  {\bf C}} required five fits, one for each fold. Because every fit
used 80\% of the development sample for training, holding the number
of HPO trials fixed resulted in approximately five times the nominal
model-training workload under {\em {\bf C}}.

Computations were performed on a home-built desktop computer running
Ubuntu 25.10, powered by a single NVIDIA GeForce RTX 5090 GPU, over
several weeks.

\subsection{Code availability}

The analysis scripts, result-level data, dataset-partitioning code,
and instructions required to reproduce the statistical analyses and
manuscript figures are provided as supplementary material. The
implementations used for image-classifier HPO with holdout
(\texttt{tunic}) and cross-validation (\texttt{cvic}) are available at
\url{https://github.com/ljbuturovic/cvic}.

\section{Results}

\begin{figure}[H]
  \begin{center}
    \includegraphics[width=0.80\linewidth]{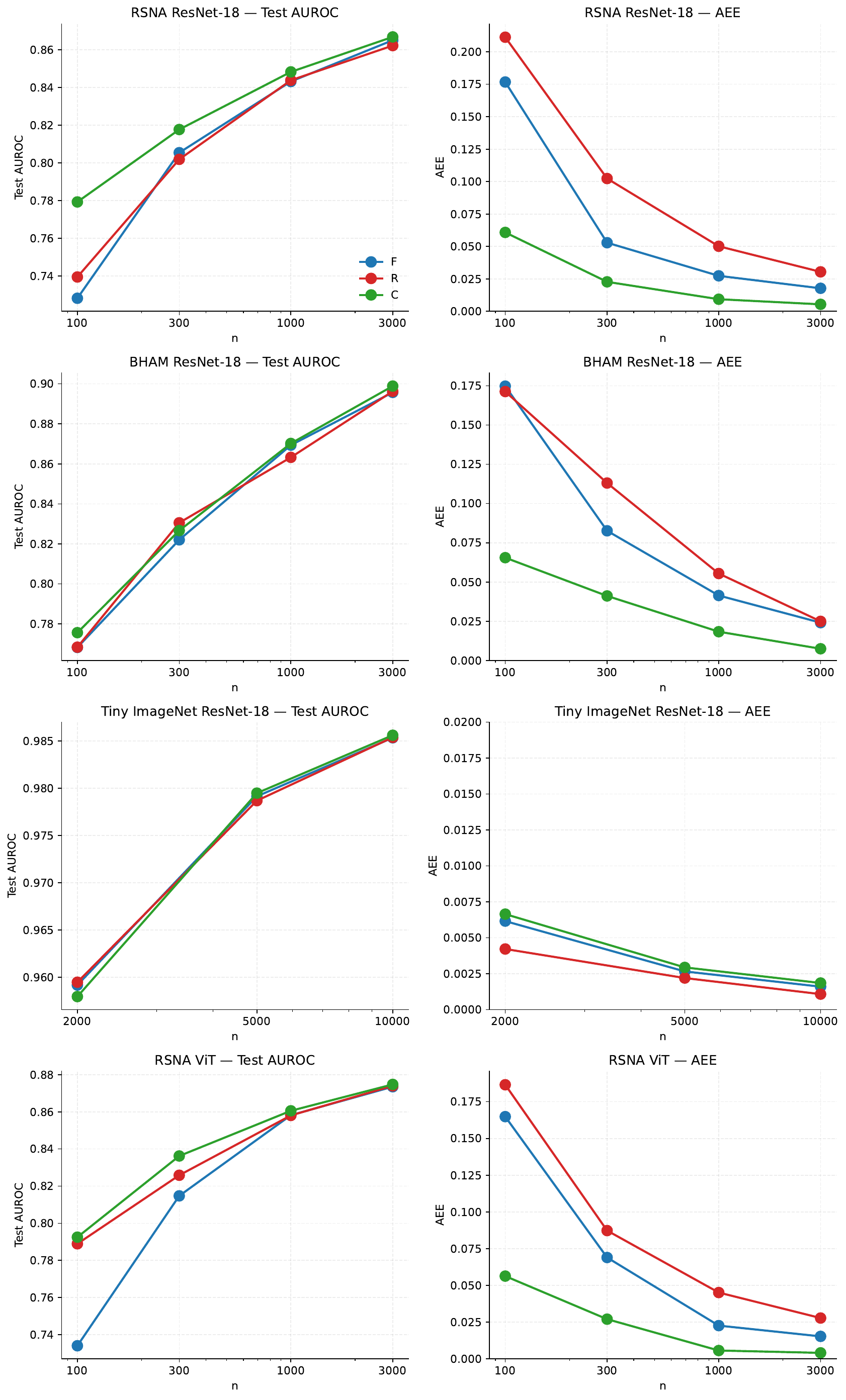}
  \end{center}
  \caption{Comparison of test set performance and absolute performance
    estimation error for different datasets, HPO methods and
    architectures. First three rows are RSNA, BHAM and TIN,
    respectively, using ResNet-18 backbone; last row is RSNA using ViT
    backbone. The x axis is the subsample size $n$ (the number of
    labeled images drawn for the run). For {\em {\bf C}}, all $n$
    images are used in cross-validation; for {\em {\bf F}}/{\em {\bf
        R}}, $n$ is split 80/20 into training and validation, so the
    plotted value $n$ includes the held-out validation images (the
    model trains on $0.8\,n$).  The y axis is test set AUROC in the
    first column and AEE in the second. Points show means across
    subsamples and random seeds; lines are included to show trends
    with development sample size. Formal paired comparisons between
    protocols are reported in Table~\ref{tab:paired-differences} and
    in the text.}
  \label{fig1}
\end{figure}

Figure~\ref{fig1} presents descriptive protocol-specific means for
test AUROC and AEE across all experimental conditions.
Table~\ref{tab:paired-differences} reports the primary paired AEE
comparisons between cross-validation and the two holdout protocols
across all medical-dataset, architecture, and sample-size conditions.
Throughout, protocol differences were calculated within paired
(subsample, seed) cells. Unless otherwise stated, reported 95\%
confidence intervals (CIs) were obtained by parametric bootstrap from
crossed random-effects models with random intercepts for development
subsample and seed.

Complete primary pairwise estimates for AEE and test AUROC across all
experimental conditions are reported in
Tables~\ref{tab:all-pairwise-A1} and
\ref{tab:all-pairwise-A2}. Sensitivity analyses were concordant for
all cross-validation-versus-holdout AEE comparisons and for 44 of 45
AEE comparisons overall. Test-AUROC intervals were less stable across
methods, but this did not alter the conclusion that differences in
final test performance were generally small and inconsistent
(Table~\ref{tab:sensitivity-agreement}).

\begin{table}[t]

  \caption{Paired differences in absolute estimation error between
    cross-validation and holdout protocols on the medical
    datasets. Differences are calculated as first protocol minus
    second protocol; negative values therefore favor the first
    protocol by indicating lower AEE. Values are mean paired
    differences with pointwise 95\% confidence intervals from crossed
    random-effects models with random intercepts for development
    subsample and seed; confidence intervals were obtained by
    parametric bootstrap with 10,000 replicates. Each comparison
    comprises 25 paired runs (five subsamples crossed with five
    seeds).}

  \label{tab:paired-differences}

  \begin{center}
    \small
    \setlength{\tabcolsep}{4pt}
    \begin{tabular}{cccc}
      \multicolumn{1}{c}{\bf Dataset}
      & \multicolumn{1}{c}{\bf $n$}
      & \multicolumn{1}{c}{\bf Comparison}
      & \multicolumn{1}{c}{\bf $\Delta$ AEE}
      \\ \hline \\

      RSNA (ResNet-18)
      & 100
      & {\bf C} $-$ {\bf F}
      & $-0.116\,[-0.152,-0.081]$
      \\
      
      & 
      & {\bf C} $-$ {\bf R}
      & $-0.150\,[-0.186,-0.116]$
      \\
      
      & 300
      & {\bf C} $-$ {\bf F}
      & $-0.030\,[-0.068,+0.008]$
      \\
      
      & 
      & {\bf C} $-$ {\bf R}
      & $-0.080\,[-0.104,-0.054]$
      \\
      
      & 1000
      & {\bf C} $-$ {\bf F}
      & $-0.018\,[-0.032,-0.003]$
      \\
      
      & 
      & {\bf C} $-$ {\bf R}
      & $-0.041\,[-0.050,-0.032]$
      \\
      
      & 3000
      & {\bf C} $-$ {\bf F}
      & $-0.012\,[-0.018,-0.007]$
      \\
      
      & 
      & {\bf C} $-$ {\bf R}
      & $-0.025\,[-0.030,-0.020]$
      \\
      \hline

      BHAM (ResNet-18)
      & 100
      & {\bf C} $-$ {\bf F}
      & $-0.109\,[-0.158,-0.061]$
      \\
      
      & 
      & {\bf C} $-$ {\bf R}
      & $-0.106\,[-0.135,-0.077]$
      \\
      
      & 300
      & {\bf C} $-$ {\bf F}
      & $-0.041\,[-0.055,-0.028]$
      \\
      
      & 
      & {\bf C} $-$ {\bf R}
      & $-0.072\,[-0.085,-0.059]$
      \\
      
      & 1000
      & {\bf C} $-$ {\bf F}
      & $-0.023\,[-0.039,-0.007]$
      \\
      
      & 
      & {\bf C} $-$ {\bf R}
      & $-0.037\,[-0.045,-0.029]$
      \\
      
      & 3000
      & {\bf C} $-$ {\bf F}
      & $-0.017\,[-0.022,-0.011]$
      \\
      
      & 
      & {\bf C} $-$ {\bf R}
      & $-0.017\,[-0.021,-0.014]$
      \\
      \hline

      RSNA (ViT)
      & 100
      & {\bf C} $-$ {\bf F}
      & $-0.109\,[-0.164,-0.053]$
      \\
      
      & 
      & {\bf C} $-$ {\bf R}
      & $-0.130\,[-0.152,-0.109]$
      \\
      
      & 300
      & {\bf C} $-$ {\bf F}
      & $-0.042\,[-0.060,-0.023]$
      \\
      
      & 
      & {\bf C} $-$ {\bf R}
      & $-0.060\,[-0.099,-0.023]$
      \\
      
      & 1000
      & {\bf C} $-$ {\bf F}
      & $-0.017\,[-0.031,-0.002]$
      \\
      
      & 
      & {\bf C} $-$ {\bf R}
      & $-0.039\,[-0.049,-0.030]$
      \\
      
      & 3000
      & {\bf C} $-$ {\bf F}
      & $-0.011\,[-0.016,-0.007]$
      \\
      
      & 
      & {\bf C} $-$ {\bf R}
      & $-0.024\,[-0.027,-0.020]$
      \\

    \end{tabular}
  \end{center}
\end{table}

\subsection{Cross-validation reduces performance-estimation error on the medical datasets}

Absolute estimation error was lower under {\em {\bf C}} than under
either holdout protocol in every medical-dataset, architecture, and
sample-size condition (Fig.~\ref{fig1}). Across the 24
cross-validation-versus-holdout comparisons in
Table~\ref{tab:paired-differences}, every point estimate favored {\em {\bf C}}.
The reductions in AEE were largest at small development
sample sizes and generally diminished as $n$ increased. The confidence
intervals reported below are pointwise 95\% intervals; multiplicity
across the 24 primary comparisons is addressed separately below.

On RSNA with ResNet-18, the {\bf C} $-$ {\bf R} difference in AEE
(negative values indicate lower error under {\em {\bf C}}) declined in
magnitude from $-0.150$ [$-0.186$, $-0.116$] at $n=100$ to $-0.025$
[$-0.030$, $-0.020$] at $n=3000$. The corresponding {\bf C} $-$ {\bf
  F} difference declined from $-0.116$ [$-0.152$, $-0.081$] to
$-0.012$ [$-0.018$, $-0.007$].

BHAM showed the same pattern. At $n=100$, the {\bf C} $-$ {\bf F}
difference in AEE was $-0.109$ [$-0.158$, $-0.061$], and the {\bf C}
$-$ {\bf R} difference was $-0.106$ [$-0.135$, $-0.077$].  The
differences decreased with increasing sample size. At $n=3000$, the
{\bf C} $-$ {\bf F} and {\bf C} $-$ {\bf R} differences were $-0.017$
[$-0.022$, $-0.011$] and $-0.017$ [$-0.021$, $-0.014$], respectively.

The RSNA analysis using the ViT backbone reproduced the ResNet-18
pattern. The {\bf C} $-$ {\bf F} difference declined in magnitude from
$-0.109$ [$-0.164$, $-0.053$] at $n=100$ to $-0.011$ [$-0.016$,
  $-0.007$] at $n=3000$. The corresponding {\bf C} $-$ {\bf R}
difference declined from $-0.130$ [$-0.152$, $-0.109$] to $-0.024$
[$-0.027$, $-0.020$].
The overall pattern remained robust after adjustment for multiplicity.
Using Bonferroni-adjusted simultaneous confidence intervals
controlling the family-wise error rate at 0.05 across the 24 primary
comparisons, all 12 {\bf C} $-$ {\bf R} intervals and 8 of 12 {\bf C}
$-$ {\bf F} intervals remained entirely below zero. Among the four
adjusted {\bf C} $-$ {\bf F} intervals that included zero, three had
upper limits no greater than $+0.005$ AUROC; thus, they permitted at
most a negligible disadvantage to {\em {\bf C}} while remaining
compatible with appreciably larger reductions in AEE. Only the RSNA
ResNet-18 comparison at $n=300$ retained substantial uncertainty about
direction.

AEE decreased with development sample size under all three protocols,
indicating that the winning configuration's validation AUROC became a
more accurate predictor of subsequent test performance as more data
became available. For medical datasets, cross-validation's absolute
advantage therefore became smaller at larger $n$.

On TIN images, AEE was negligible under all three protocols at every
sample size. The largest pairwise AEE difference was 0.003 AUROC, and
every CI included zero. Thus, there was no practically meaningful
performance-estimation advantage for any protocol on this data.

\subsection{Fixed holdout tends to have lower absolute estimation error than reshuffled holdout}

Comparing the two holdout protocols, {\em {\bf F}} had lower mean AEE
than {\em {\bf R}} in 11 of the 12 medical-dataset, architecture, and
sample-size conditions (Table~\ref{tab:fixed-reshuffled}). In the
primary crossed random-effects analysis, the {\bf F} $-$ {\bf R}
confidence interval excluded zero in 6 of the 12 conditions. The sole
point estimate favoring {\em {\bf R}} occurred on BHAM at $n=100$ and
was close to zero ({\bf F} $-$ {\bf R}: $+0.003$
[$-0.060$, $+0.066$]).

\begin{table}[t]

  \caption{Paired differences in absolute estimation error between
    fixed and reshuffled holdout on the medical datasets. Differences
    are calculated as {\bf F} minus {\bf R}; negative values therefore
    favor fixed holdout by indicating lower AEE. Values are mean
    paired differences with pointwise 95\% confidence intervals from
    crossed random-effects models with random intercepts for
    development subsample and seed; confidence intervals were obtained
    by parametric bootstrap with 10,000 replicates. Each comparison
    comprises 25 paired runs (five subsamples crossed with five
    seeds).}

  \label{tab:fixed-reshuffled}

  \begin{center}
    \begin{tabular}{ccc}
      \multicolumn{1}{c}{\bf Dataset}
      & \multicolumn{1}{c}{\bf $n$}
      & \multicolumn{1}{c}{\bf $\Delta$ AEE: {\bf F} $-$ {\bf R}}
      \\ \hline \\

      RSNA (ResNet-18)
      & 100
      & $-0.035\,[-0.076,+0.008]$
      \\
      
      & 300
      & $-0.049\,[-0.080,-0.018]$
      \\
      
      & 1000
      & $-0.023\,[-0.040,-0.005]$
      \\
      
      & 3000
      & $-0.013\,[-0.020,-0.005]$
      \\
      \\

      BHAM (ResNet-18)
      & 100
      & $+0.003\,[-0.060,+0.066]$
      \\
      
      & 300
      & $-0.030\,[-0.045,-0.016]$
      \\
      
      & 1000
      & $-0.014\,[-0.030,+0.002]$
      \\
      
      & 3000
      & $-0.001\,[-0.006,+0.005]$
      \\
      \\

      RSNA (ViT)
      & 100
      & $-0.022\,[-0.095,+0.049]$
      \\
      
      & 300
      & $-0.018\,[-0.057,+0.019]$
      \\
      
      & 1000
      & $-0.023\,[-0.036,-0.008]$
      \\
      
      & 3000
      & $-0.012\,[-0.019,-0.006]$
      \\
    \end{tabular}
  \end{center}
\end{table}

On RSNA with ResNet-18, the {\bf F} $-$ {\bf R} point estimate was
negative at every sample size. The difference was $-0.035$
[$-0.076$, $+0.008$] at $n=100$ and $-0.049$
[$-0.080$, $-0.018$] at $n=300$. It remained negative at $n=1000$
and $n=3000$, with estimates of $-0.023$
[$-0.040$, $-0.005$] and $-0.013$ [$-0.020$, $-0.005$],
respectively. Thus, the primary CI excluded zero at every sample size
except $n=100$.

On BHAM, the protocols had similar AEE at $n=100$. At $n=300$,
{\em {\bf F}} had lower AEE than {\em {\bf R}} by 0.030 AUROC
({\bf F} $-$ {\bf R}: $-0.030$ [$-0.045$, $-0.016$]). The point
estimates also favored {\em {\bf F}} at $n=1000$ and $n=3000$, but
the differences were smaller and their CIs included zero:
$-0.014$ [$-0.030$, $+0.002$] and $-0.001$
[$-0.006$, $+0.005$], respectively.

The RSNA ViT analysis showed the same general tendency. The
{\bf F} $-$ {\bf R} point estimate was negative at every sample size,
although the CI excluded zero only at $n=1000$
($-0.023$ [$-0.036$, $-0.008$]) and $n=3000$
($-0.012$ [$-0.019$, $-0.006$]). At $n=100$ and $n=300$, the
estimates were $-0.022$ [$-0.095$, $+0.049$] and $-0.018$
[$-0.057$, $+0.019$], respectively.

On Tiny ImageNet, the {\bf F} $-$ {\bf R} point estimates slightly
favored {\em {\bf R}}, but no difference exceeded 0.002 AUROC and
every CI included zero. Overall, reshuffling provided no consistent
AEE advantage and generally produced less accurate development-time
performance estimates than fixed holdout in the medical experiments.

\subsection{Test AUROC}

Differences in test AUROC were generally smaller and less consistent
than differences in AEE (Fig.~\ref{fig1};
Tables~\ref{tab:all-pairwise-A1} and \ref{tab:all-pairwise-A2}). On
RSNA with ResNet-18, {\em {\bf C}} had higher mean test AUROC than
both holdout protocols at every sample size. At $n=100$, the {\bf C}
$-$ {\bf F} difference was $+0.051$ [$+0.014$, $+0.088$], and the {\bf
  C} $-$ {\bf R} difference was $+0.040$ [$+0.008$, $+0.071$]. At
$n=300$, the corresponding differences were smaller: $+0.012$
[$+0.001$, $+0.023$] and $+0.016$ [$+0.008$, $+0.024$],
respectively. At $n=1000$ and $n=3000$, no {\em {\bf
    C}}-versus-holdout difference exceeded 0.005 AUROC.

On BHAM, protocol differences in test AUROC were small throughout. No
pairwise difference exceeded 0.009 AUROC. Most primary CIs included
zero; the exceptions were the {\bf C} $-$ {\bf R} comparison at
$n=1000$ ($+0.007$ [$+0.001$, $+0.013$]) and the {\bf C} $-$ {\bf F}
comparison at $n=3000$, where the estimated advantage was only 0.003
AUROC. Thus, the large reductions in AEE observed on BHAM were not
accompanied by comparably large improvements in final test
performance.

The RSNA ViT results were more variable at the smallest sample size.
At $n=100$, the {\bf C} $-$ {\bf F} point estimate was $+0.058$ and
the {\bf F} $-$ {\bf R} estimate was $-0.055$, but the primary CIs for
both comparisons included zero ($[-0.021$, $+0.138$] and $[-0.143$,
  $+0.035$], respectively). At $n=300$, {\em {\bf C}} exceeded {\em
  {\bf F}} by 0.021 AUROC [$+0.0003$, $+0.042$] and {\em {\bf R}} by
0.010 AUROC [$+0.001$, $+0.019$]. At $n=1000$ and $n=3000$, every
pairwise difference was at most 0.0025 AUROC and every primary CI
included zero.

On Tiny ImageNet, pairwise differences in test AUROC were negligible:
none exceeded 0.002 AUROC in magnitude. Although several primary CIs
narrowly excluded zero, the estimated differences were too small to be
practically meaningful.

Thus, the clear and consistent advantage of cross-validation in AEE
did not translate into a similarly consistent advantage in final test
AUROC. Test performance was generally similar among the protocols,
with modest advantages for {\em {\bf C}} confined primarily to several
RSNA conditions.

\subsection{Robustness to backbone architecture}

The primary AEE findings on RSNA were closely reproduced when the
ResNet-18 backbone was replaced with ViT-S/16. Under both
architectures, {\em {\bf C}} had lower mean AEE than {\em {\bf F}} and
{\em {\bf R}} at every sample size, and the magnitude of the
differences decreased as $n$ increased
(Table~\ref{tab:paired-differences}). The effect sizes were also
similar across architectures. For example, the {\bf C} $-$ {\bf F}
difference declined from $-0.116$ to $-0.012$ between $n=100$ and
$n=3000$ with ResNet-18, compared with $-0.109$ to $-0.011$ with
ViT-S/16. The corresponding {\bf C} $-$ {\bf R} differences declined
from $-0.150$ to $-0.025$ with ResNet-18 and from $-0.130$ to $-0.024$
with ViT-S/16.

The comparison between the two holdout protocols was directionally
consistent across backbones: {\em {\bf F}} had lower mean AEE than
{\em {\bf R}} at every RSNA sample size under both ResNet-18 and
ViT-S/16. Under ViT-S/16, the {\bf F} $-$ {\bf R} CI excluded zero at
$n=1000$ and $n=3000$, matching the tendency observed with ResNet-18.

Test-AUROC results were less uniform. Under both architectures, {\em
  {\bf C}} showed its largest potential advantage at the smaller
sample sizes, while pairwise differences were negligible by $n=1000$
and $n=3000$. The exact protocol ordering at $n=100$ differed between
architectures: with ViT-S/16, {\em {\bf R}} had higher mean test AUROC
than {\em {\bf F}}, whereas the corresponding ResNet-18 difference was
small. Thus, the reduction in AEE under cross-validation was robust to
the change in backbone architecture, whereas small differences in
final test performance were not.

\subsection{Tiny ImageNet top-1 accuracy}

Because macro-averaged one-vs-rest AUROC does not directly measure
exact class identification in a 200-class problem, we additionally
evaluated top-1 accuracy for the Tiny ImageNet models selected using
AUROC. Top-1 accuracy was a secondary evaluation metric and played no
role in HPO or model selection.

\begin{figure}[H]
  \begin{center}
    \includegraphics[width=0.65\linewidth]{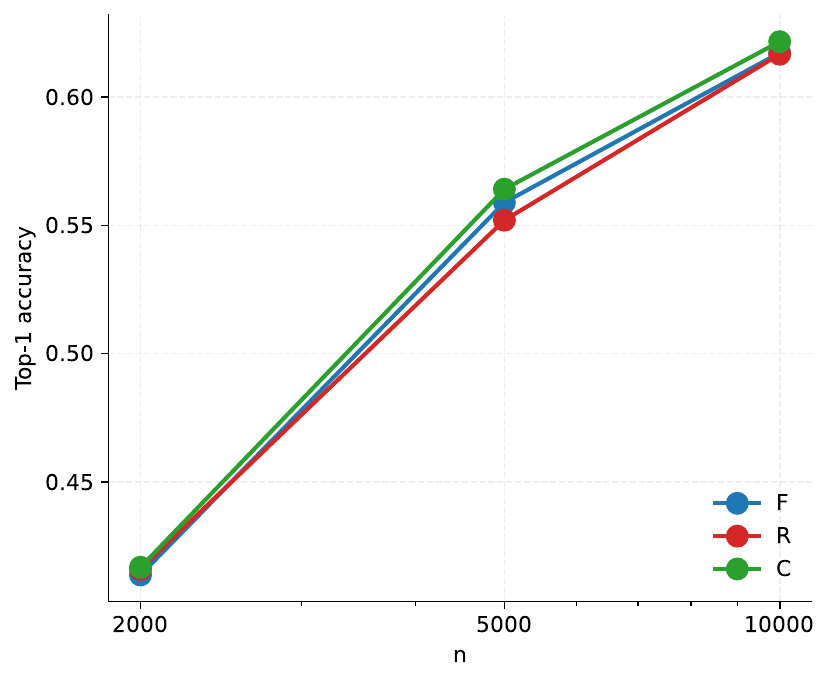}
  \end{center}
  \caption{Comparison of top-1 accuracy for TIN dataset for different
    sample sizes.}
  \label{fig:tin-top1}
\end{figure}

Mean top-1 accuracy increased substantially with development sample
size under all three protocols (Fig.~\ref{fig:tin-top1}). At $n=2000$,
mean accuracy was 0.414 under {\em {\bf F}}, 0.416 under {\em {\bf
    R}}, and 0.417 under {\em {\bf C}}. At $n=5000$, the corresponding
values were 0.559, 0.552, and 0.564, and at $n=10000$ they were 0.617,
0.617, and 0.622. Thus, accuracy increased from approximately 0.41 to
0.62 as $n$ increased, confirming that the task was not saturated
despite the high macro-AUROC values.

Differences among protocols were small relative to the improvement
associated with increasing sample size. At $n=2000$, no paired
difference exceeded 0.003. {\em {\bf C}} had the highest mean accuracy
at $n=5000$ and $n=10000$, but its largest advantage was 0.012,
relative to {\em {\bf R}} at $n=5000$. Overall, protocol choice had
little effect on top-1 accuracy, consistent with the finding that
differences in Tiny ImageNet test AUROC were also negligible.

\subsection{Sensitivity analyses}

We evaluated the robustness of the primary crossed random-effects
analysis using a two-way cluster bootstrap over subsamples and seeds
and an analysis based on seed-averaged subsample differences. All
three analyses estimated the same paired mean differences and differed
only in their construction of the pointwise 95\% confidence intervals.

\begin{table}[t]
  \caption{Agreement between the primary and sensitivity analyses. A
    comparison was classified as concordant when all three pointwise
    95\% confidence intervals either included zero or excluded zero in
    the same direction. Each row comprises 15 comparisons across the
    applicable datasets, architectures, and sample sizes.}
  \label{tab:sensitivity-agreement}
  \begin{center}
    \begin{tabular}{llcc}
      \multicolumn{1}{c}{\bf Metric}
      & \multicolumn{1}{c}{\bf Comparison}
      & \multicolumn{1}{c}{\bf Concordant}
      & \multicolumn{1}{c}{\bf Total}
      \\ \hline \\
      AEE        & {\bf C} $-$ {\bf F} & 15 & 15 \\
      AEE        & {\bf C} $-$ {\bf R} & 15 & 15 \\
      AEE        & {\bf F} $-$ {\bf R} & 14 & 15 \\
      Test AUROC & {\bf C} $-$ {\bf F} &  9 & 15 \\
      Test AUROC & {\bf C} $-$ {\bf R} &  9 & 15 \\
      Test AUROC & {\bf F} $-$ {\bf R} & 13 & 15 \\
    \end{tabular}
  \end{center}
\end{table}

The AEE results were highly robust to the method used to estimate
uncertainty. The three analyses were concordant in 44 of 45
comparisons, including all 30 comparisons between {\em {\bf C}} and a
holdout protocol (Table~\ref{tab:sensitivity-agreement}). The sole
discrepancy was the {\bf F} $-$ {\bf R} comparison on RSNA with
ResNet-18 at $n=1000$. The point estimate was $-0.023$ under all three
analyses; the crossed-model CI was $[-0.040,-0.005]$, the two-way
cluster-bootstrap CI was $[-0.040,-0.003]$, and the CI based on
seed-averaged subsample differences was $[-0.047,+0.002]$.  Thus, this
discrepancy concerned whether a single interval narrowly included zero
and did not affect the direction or magnitude of the estimated
difference.

For test AUROC, the analyses were concordant in 31 of 45 comparisons.
All 14 discrepancies concerned whether an interval included zero; no
comparison yielded confidence intervals excluding zero in opposite
directions. Four discrepancies involved the larger
cross-validation-versus-holdout estimates on RSNA at small sample
sizes: the ResNet-18 comparisons at $n=100$ and the ViT comparisons at
$n=300$. In each case, the crossed-model and two-way cluster-bootstrap
intervals excluded zero, whereas the interval based on five
seed-averaged subsample values included zero. The remaining
discrepancies involved point estimates no larger than 0.007 AUROC in
magnitude. Accordingly, variation among the interval estimators did
not alter the conclusion that differences in final test performance
were generally small and less consistent than the differences in AEE.

\section{Discussion}

Cross-validation consistently reduced absolute performance-estimation
error relative to holdout methods on the medical datasets, with the
largest advantage at small sample sizes and a diminishing advantage as
$n$ increased.  This result was robust to conservative family-wise
adjustment: all 24 point estimates favored cross-validation, and most
simultaneous confidence intervals remained entirely below zero. For
the TIN dataset, AEE was practically zero under every HPO approach,
hence there was little for CV to improve. These reductions in AEE did
not generally translate into higher test AUROC, which was largely
similar among protocols. We conclude that, for small sample sizes,
holdout validation can provide an inaccurate estimate of subsequent
test performance, whereas CV can make that estimate substantially more
reliable.

We found that the AEE advantage of CV was not specific to the
ResNet-18 backbone, since we observed virtually the same effect using
ViT-S/16.

In secondary analysis, we found that reshuffling the validation split
for each trial, which has been reported to improve generalization for
non-DL models, generally increased AEE in our deep
image-classification setting. Thus, we found no empirical
justification for reshuffling rather than reusing a fixed holdout in
this context.

Five-fold {\em {\bf C}} required approximately five times the nominal
model-training workload of {\em {\bf F}} or {\em {\bf R}}, because
each HPO trial required five model fits rather than one. We did not
measure wall-clock overhead in a controlled setting; the actual
elapsed-time cost will depend on the hardware, implementation, and
degree of parallel execution.

Not every development scenario conforms to the setup studied here (a
fixed development pool for HPO followed by a single final test
evaluation). Some workflows reserve an additional holdout set during
development to detect overfitting or other failures. We acknowledge we
are unable to anticipate or analyze every conceivable development
scenario.  Nevertheless, the key point is the overall pool of samples
available for development. We believe there is no escaping the impact
of a small sample size simply by shifting images among data
partitions. In particular, setting aside an additional holdout set
means fewer images are available for training and validation, thereby
reducing the data available for learning neural network weights. When
acquiring more data is prohibitively expensive, computationally
intensive cross-validation can provide a more reliable validation
signal while allowing every development image to contribute to both
training and validation across folds. Therefore, instead of further
dividing a limited dataset, we propose using all available development
data within cross-validation for HPO and then retraining the selected
configuration on the full development set. In our experiments, this
approach substantially reduced performance-estimation error by trading
computational resources for a more reliable estimate of subsequent
test performance.

\subsection{Limitations}

The entire analysis matrix comprised 300 final models for BHAM, 600
for RSNA, and 135 for TIN. Within each dataset, all models were
evaluated on the same held-out test set. The resulting
test-performance estimates were therefore correlated because they were
calculated on the same cases; they should not be interpreted as
hundreds of independent evaluations. This reuse does not by itself
imply test-set overfitting: the experimental runs were predefined and
automated, and test performance was not used to select
hyperparameters, models, protocols, or subsequent experimental
decisions. Nevertheless, all estimates remain conditional on the same
finite test samples. A strictly independent assessment of every final
model would require a separate, previously unseen dataset (such as a
prospectively collected external or pivotal validation cohort) for
{\em each} model. This is not practically feasible.

Computational constraints limited the design to five development
subsamples per dataset and sample size. This limits the precision with
which variability across possible development samples can be
estimated. We addressed the crossed reuse of subsamples and random
seeds using crossed random-effects models and complementary
sensitivity analyses, which produced consistent conclusions for the
primary AEE comparisons. Additional independently drawn subsamples
would nevertheless permit more precise estimation.

Previous research recommended repeated CV~\citep{krstajic2014cross}
for HPO. Repeating the full cross-validation procedure was infeasible
with our available computational resources and is left for future
research. The single five-fold CV procedure evaluated in this
manuscript nevertheless substantially reduced AEE across the medical
experiments.

\section{Conclusions}

Cross-validation substantially reduces absolute performance-estimation
error on small medical image datasets, making development-time
performance estimates more reliable. This conclusion was robust to
conservative family-wise adjustment. The advantage diminishes as
sample size increases and disappears when estimation error is already
negligible. Final test performance is generally similar between
cross-validation and holdout protocols.

We therefore recommend cross-validation-based HPO for small-sample
deep image classification when computational resources permit. It
trades additional computation for a more reliable estimate of
subsequent test performance.

As a secondary finding, fixed holdout tends to provide more accurate
performance estimates than reshuffled holdout, although this result
was less uniformly supported.

\bibliography{main}
\bibliographystyle{tmlr}

\appendix

\section{Appendix}

\begin{table}[p]

  \caption{Complete paired comparisons of absolute estimation error
    and test AUROC across all experimental conditions, for RSNA and
    BHAM, using ResNet-18. Differences are calculated as the first
    protocol minus the second protocol. Negative values in $\Delta$
    AEE favor the first protocol because they indicate lower absolute
    estimation error; positive values in $\Delta$ test AUROC favor the
    first protocol. Values are mean paired differences with pointwise
    95\% confidence intervals from crossed random-effects models with
    random intercepts for development subsample and seed; confidence
    intervals were obtained by parametric bootstrap with 10,000
    replicates. The comparisons comprise 25 paired runs (five
    subsamples crossed with five seeds).}

  \label{tab:all-pairwise-A1}

  \begin{center}
    \small
    \begin{tabular}{ccccc}
      \multicolumn{1}{c}{\bf Dataset}
      & \multicolumn{1}{c}{\bf $n$}
      & \multicolumn{1}{c}{\bf Comparison}
      & \multicolumn{1}{c}{\bf $\Delta$ AEE}
      & \multicolumn{1}{c}{\bf $\Delta$ test AUROC}
      \\ \hline \\

      RSNA (ResNet-18)
      & 100
      & {\bf C} $-$ {\bf F}
      & $-0.116\,[-0.152,-0.081]$
      & $+0.051\,[+0.014,+0.088]$
      \\
      
      & 
      & {\bf C} $-$ {\bf R}
      & $-0.150\,[-0.186,-0.116]$
      & $+0.040\,[+0.008,+0.071]$
      \\
      
      & 
      & {\bf F} $-$ {\bf R}
      & $-0.035\,[-0.076,+0.008]$
      & $-0.011\,[-0.067,+0.045]$
      \\
      
      & 300
      & {\bf C} $-$ {\bf F}
      & $-0.030\,[-0.068,+0.008]$
      & $+0.012\,[+0.001,+0.023]$
      \\
      
      & 
      & {\bf C} $-$ {\bf R}
      & $-0.080\,[-0.104,-0.054]$
      & $+0.016\,[+0.008,+0.024]$
      \\
      
      & 
      & {\bf F} $-$ {\bf R}
      & $-0.049\,[-0.080,-0.018]$
      & $+0.004\,[-0.009,+0.017]$
      \\
      
      & 1000
      & {\bf C} $-$ {\bf F}
      & $-0.018\,[-0.032,-0.003]$
      & $+0.005\,[+0.002,+0.008]$
      \\
      
      & 
      & {\bf C} $-$ {\bf R}
      & $-0.041\,[-0.050,-0.032]$
      & $+0.004\,[-0.001,+0.010]$
      \\
      
      & 
      & {\bf F} $-$ {\bf R}
      & $-0.023\,[-0.040,-0.005]$
      & $-0.001\,[-0.005,+0.004]$
      \\
      
      & 3000
      & {\bf C} $-$ {\bf F}
      & $-0.012\,[-0.018,-0.007]$
      & $+0.002\,[+0.001,+0.003]$
      \\
      
      & 
      & {\bf C} $-$ {\bf R}
      & $-0.025\,[-0.030,-0.020]$
      & $+0.005\,[+0.003,+0.007]$
      \\
      
      & 
      & {\bf F} $-$ {\bf R}
      & $-0.013\,[-0.020,-0.005]$
      & $+0.003\,[+0.001,+0.005]$
      \\
      \\

      BHAM (ResNet-18)
      & 100
      & {\bf C} $-$ {\bf F}
      & $-0.109\,[-0.158,-0.061]$
      & $+0.007\,[-0.034,+0.047]$
      \\
      
      & 
      & {\bf C} $-$ {\bf R}
      & $-0.106\,[-0.135,-0.077]$
      & $+0.007\,[-0.024,+0.039]$
      \\
      
      & 
      & {\bf F} $-$ {\bf R}
      & $+0.003\,[-0.060,+0.066]$
      & $-0.000\,[-0.058,+0.058]$
      \\
      
      & 300
      & {\bf C} $-$ {\bf F}
      & $-0.041\,[-0.055,-0.028]$
      & $+0.005\,[-0.005,+0.014]$
      \\
      
      & 
      & {\bf C} $-$ {\bf R}
      & $-0.072\,[-0.085,-0.059]$
      & $-0.004\,[-0.013,+0.005]$
      \\
      
      & 
      & {\bf F} $-$ {\bf R}
      & $-0.030\,[-0.045,-0.016]$
      & $-0.009\,[-0.018,+0.000]$
      \\
      
      & 1000
      & {\bf C} $-$ {\bf F}
      & $-0.023\,[-0.039,-0.007]$
      & $+0.001\,[-0.004,+0.006]$
      \\
      
      & 
      & {\bf C} $-$ {\bf R}
      & $-0.037\,[-0.045,-0.029]$
      & $+0.007\,[+0.001,+0.013]$
      \\
      
      & 
      & {\bf F} $-$ {\bf R}
      & $-0.014\,[-0.030,+0.002]$
      & $+0.006\,[-0.000,+0.013]$
      \\
      
      & 3000
      & {\bf C} $-$ {\bf F}
      & $-0.017\,[-0.022,-0.011]$
      & $+0.003\,[+0.000,+0.006]$
      \\
      
      & 
      & {\bf C} $-$ {\bf R}
      & $-0.017\,[-0.021,-0.014]$
      & $+0.003\,[-0.000,+0.006]$
      \\
      
      & 
      & {\bf F} $-$ {\bf R}
      & $-0.001\,[-0.006,+0.005]$
      & $-0.001\,[-0.004,+0.003]$
      \\
    \end{tabular}
  \end{center}
\end{table}

\begin{table}[p]

  \caption{Complete paired comparisons of absolute estimation error
    and test AUROC across all experimental conditions for RSNA using
    ViT architecture, and TIN. Differences are calculated as the first
    protocol minus the second protocol. Negative values in $\Delta$
    AEE favor the first protocol because they indicate lower absolute
    estimation error; positive values in $\Delta$ test AUROC favor the
    first protocol. Values are mean paired differences with pointwise
    95\% confidence intervals from crossed random-effects models with
    random intercepts for development subsample and seed; confidence
    intervals were obtained by parametric bootstrap with 10,000
    replicates. RSNA comparisons comprise 25 paired runs (five
    subsamples crossed with five seeds), and Tiny ImageNet comparisons
    comprise 15 paired runs (five subsamples crossed with three
    seeds).}

  \label{tab:all-pairwise-A2}

  \begin{center}
    \small
    \begin{tabular}{ccccc}
      \multicolumn{1}{c}{\bf Dataset}
      & \multicolumn{1}{c}{\bf $n$}
      & \multicolumn{1}{c}{\bf Comparison}
      & \multicolumn{1}{c}{\bf $\Delta$ AEE}
      & \multicolumn{1}{c}{\bf $\Delta$ test AUROC}
      \\ \hline \\

      RSNA (ViT)
      & 100
      & {\bf C} $-$ {\bf F}
      & $-0.109\,[-0.164,-0.053]$
      & $+0.058\,[-0.021,+0.138]$
      \\
      
      & 
      & {\bf C} $-$ {\bf R}
      & $-0.130\,[-0.152,-0.109]$
      & $+0.004\,[-0.013,+0.020]$
      \\
      
      & 
      & {\bf F} $-$ {\bf R}
      & $-0.022\,[-0.095,+0.049]$
      & $-0.055\,[-0.143,+0.035]$
      \\
      
      & 300
      & {\bf C} $-$ {\bf F}
      & $-0.042\,[-0.060,-0.023]$
      & $+0.021\,[+0.000,+0.042]$
      \\
      
      & 
      & {\bf C} $-$ {\bf R}
      & $-0.060\,[-0.099,-0.023]$
      & $+0.010\,[+0.001,+0.019]$
      \\
      
      & 
      & {\bf F} $-$ {\bf R}
      & $-0.018\,[-0.057,+0.019]$
      & $-0.011\,[-0.035,+0.015]$
      \\
      
      & 1000
      & {\bf C} $-$ {\bf F}
      & $-0.017\,[-0.031,-0.002]$
      & $+0.002\,[-0.001,+0.005]$
      \\
      
      & 
      & {\bf C} $-$ {\bf R}
      & $-0.039\,[-0.049,-0.030]$
      & $+0.002\,[-0.001,+0.006]$
      \\
      
      & 
      & {\bf F} $-$ {\bf R}
      & $-0.023\,[-0.036,-0.008]$
      & $+0.000\,[-0.004,+0.004]$
      \\
      
      & 3000
      & {\bf C} $-$ {\bf F}
      & $-0.011\,[-0.016,-0.007]$
      & $+0.001\,[-0.000,+0.003]$
      \\
      
      & 
      & {\bf C} $-$ {\bf R}
      & $-0.024\,[-0.027,-0.020]$
      & $+0.001\,[-0.001,+0.002]$
      \\
      
      & 
      & {\bf F} $-$ {\bf R}
      & $-0.012\,[-0.019,-0.006]$
      & $-0.000\,[-0.003,+0.002]$
      \\
      \\

      Tiny ImageNet
      & 2000
      & {\bf C} $-$ {\bf F}
      & $+0.000\,[-0.002,+0.003]$
      & $-0.001\,[-0.004,+0.002]$
      \\
      
      & 
      & {\bf C} $-$ {\bf R}
      & $+0.002\,[-0.001,+0.006]$
      & $-0.002\,[-0.004,+0.001]$
      \\
      
      & 
      & {\bf F} $-$ {\bf R}
      & $+0.002\,[-0.001,+0.005]$
      & $-0.000\,[-0.002,+0.001]$
      \\
      
      & 5000
      & {\bf C} $-$ {\bf F}
      & $+0.000\,[-0.001,+0.001]$
      & $+0.000\,[-0.000,+0.001]$
      \\
      
      & 
      & {\bf C} $-$ {\bf R}
      & $+0.001\,[-0.000,+0.002]$
      & $+0.001\,[+0.000,+0.002]$
      \\
      
      & 
      & {\bf F} $-$ {\bf R}
      & $+0.000\,[-0.001,+0.001]$
      & $+0.000\,[+0.000,+0.001]$
      \\
      
      & 10000
      & {\bf C} $-$ {\bf F}
      & $+0.000\,[-0.001,+0.001]$
      & $+0.000\,[+0.000,+0.001]$
      \\
      
      & 
      & {\bf C} $-$ {\bf R}
      & $+0.001\,[-0.000,+0.002]$
      & $+0.000\,[-0.000,+0.001]$
      \\
      
      & 
      & {\bf F} $-$ {\bf R}
      & $+0.001\,[-0.000,+0.001]$
      & $-0.000\,[-0.000,+0.000]$
      \\
    \end{tabular}
  \end{center}
\end{table}

\end{document}